\documentclass[acmsmall]{acmart}
\AtBeginDocument{%
  \providecommand\BibTeX{{%
    \normalfont B\kern-0.5em{\scshape i\kern-0.25em b}\kern-0.8em\TeX}}}

\usepackage{latexsym}

\usepackage{booktabs}
\usepackage{multirow}
\usepackage{makecell}

\usepackage{amsmath}
\usepackage{amssymb}
\usepackage{graphicx}
\usepackage{subfigure}
\usepackage{float} 
\usepackage{enumerate}

\usepackage{CJKutf8}

\begin{document}
\setcopyright{acmcopyright}
\acmJournal{TALLIP}
\acmYear{2022} \acmVolume{1} \acmNumber{1} \acmArticle{1} \acmMonth{1} \acmPrice{15.00}\acmDOI{10.1145/3564271}
\title{General and Domain Adaptive Chinese Spelling Check with Error Consistent Pretraining}

\author{Qi Lv}
\authornote{Both authors contributed equally to this research.}
\email{20205227047@stu.suda.edu.cn}
\orcid{1234-5678-9012}
\affiliation{%
  \institution{School of Computer Science and Technology, Soochow University}
  \country{China}
}
\author{Ziqiang Cao}
\authornotemark[1]
\authornote{Corresponding author.}
\email{zqcao@suda.edu.cn}
\affiliation{%
  \institution{School of Computer Science and Technology, Institution of Artificial Intelligence, Soochow University}
  \country{China}
}
\author{Lei Geng}
\affiliation{%
  \institution{School of Computer Science and Technology, Soochow University}
  \country{China}
}
\author{Chunhui Ai}
\affiliation{%
  \institution{School of Computer Science and Technology, Soochow University}
  \country{China}
}
\author{Xu Yan}
\affiliation{%
  \institution{School of Computer Science and Technology, Soochow University}
  \country{China}
}
\author{Guohong Fu}
\email{ghfu@hotmail.com}
\affiliation{%
  \institution{School of Computer Science and Technology, Institution of Artificial Intelligence, Soochow University}
  \country{China}
}

\renewcommand{\shortauthors}{Qi and Ziqiang, et al.}

\begin{abstract}
The lack of label data is one of the significant bottlenecks for Chinese Spelling Check (CSC).
Existing researches use the automatic generation method by exploiting unlabeled data to expand the supervised corpus.
However, there is a big gap between the real input scenario and automatically generated corpus.
Thus, we develop a competitive general speller \textbf{ECSpell} which adopts the \textbf{E}rror \textbf{C}onsistent masking strategy to create data for pretraining.
This error consistency masking strategy is used to specify the error types of automatically generated sentences consistent with the real scene.
The experimental result indicates that our model outperforms previous state-of-the-art models on the general benchmark.

Moreover, spellers often work within a particular domain in real life.
Due to many uncommon domain terms, experiments on our built domain specific datasets show that general models perform terribly.
Inspired by the common practice of input methods, we propose to add an alterable user dictionary to handle the zero-shot domain adaption problem.
Specifically, we attach a \textbf{U}ser \textbf{D}ictionary guided inference module (\textbf{UD}) to a general token classification based speller.
Our experiments demonstrate that ECSpell$^{UD}$, namely ECSpell combined with UD, surpasses all the other baselines broadly, even approaching the performance on the general benchmark\footnote{https://github.com/Aopolin-Lv/ECSpell}.
\end{abstract}



\maketitle

\section{Introduction}
Chinese spelling check (CSC) \cite{DBLP:conf/acl-sighan/TsengLCC15} aims to identify and correct the misspelling characters in Chinese sentences. 
Since spelling errors may cause deviations in the semantics of sentences,
the CSC task is significant to most downstream NLP applications such as named entity recognition, machine translation, and text summarization.
CSC is usually regarded as a token classification problem \cite{hong2019faspell} as the lengths of input and output are the same.

Recently, CSC has made remarkable progress with the help of the large-scale pretrained language models like BERT \cite{bert}.
\citet{hong2019faspell} firstly modeled CSC as a BERT token classification task.
Since almost all the spelling errors are related to phonological or visual similarity \cite{liu-etal-2010-visually},
many subsequent studies incorporated the similarity knowledge into spellers.
For example, \citet{DBLP:journals/taslp/NguyenNC21} utilized glyph information while
\citet{DBLP:conf/acl/ChengXCJWWCQ20, zhang-etal-2021-correcting}, and \citet{xu-etal-2021-read} employed phonetic features.
The fusion method of these similarities has also been explored,
such as Graph Convolution Network (GCN) \cite{DBLP:conf/acl/ChengXCJWWCQ20} and multi-modal \cite{huang-etal-2021-phmospell, xu-etal-2021-read}.

One of the main challenges of utilizing supervised learning for CSC is the lack of high-quality annotated corpora.
Most previous CSC studies concentrated on the general benchmark SIGHAN \cite{DBLP:conf/acl-sighan/WuLL13, DBLP:conf/acl-sighan/YuLTC14, DBLP:conf/acl-sighan/TsengLCC15}.
To alleviate this problem, large-scale unlabelled corpus was used to enhance the spelling check ability of models.
Some works \cite{DBLP:conf/nlpcc/DuanPWZW19, DBLP:conf/emnlp/WangSLHZ18, zhang-etal-2021-correcting, liu-etal-2021-plome} adopted multimodal-based methods to build corpora automatically for weakly supervised learning.
Although some of these methods \cite{DBLP:conf/emnlp/WangSLHZ18} work well, they still lack the consideration of the consistency between the real datasets and the automatically generated corpora, which will bring a big gap.

To build a practical competitive speller, we develop a novel general speller \textbf{ECSpell} which uses the \textbf{E}rror \textbf{C}onsistent masking strategy to exploit unlabeled data for pretraining.
Previous work \cite{zhang-etal-2021-correcting, liu-etal-2021-plome} ignored the significant fact that the error source of a sentence could often be the only one due to the input method.
The proposed masking strategy uses this property to simulate the natural scene as much as possible to construct the corpus automatically.
Concretely, this masking strategy first specifies an error type of the given sentence, then masks single or continuous characters according to this type to create spelling errors.
It can be simply applied to large-scale corpus for unsupervised CSC learning.
Meanwhile, we fuse the glyph information and fine-grained phonetic features.
See details in following Section \ref{ecspell_model} and Section \ref{ecm}.

In addition, spellers often work within a specific domain.
Considering most domain terms, our experiments show that general spellers perform terribly in this case.
Many such errors can not be corrected, and some correct parts are even wrongly changed.
For example, with no medical knowledge, it is difficult to detect that “Xian Bing” means “\begin{CJK}{UTF8}{gbsn}痫病\end{CJK} (Tinea)” instead of the common word “\begin{CJK}{UTF8}{gbsn}馅饼\end{CJK} (pie)”.
Under the circumstance of the document writing, it is common for a speller to wrongly change “\begin{CJK}{UTF8}{gbsn}金字村\end{CJK} (Jinzi Village)” into “\begin{CJK}{UTF8}{gbsn}金子村\end{CJK} (Golden Village)” as the lack of the particular external terms knowledge.

It is unacceptable in time and economy to annotate data in each domain and fine-tune models individually, especially for domains with increasing new terms.
Considering the actual application, we extend the traditional CSC task to the domain adaptive CSC task.
Following the common practice of the industrial word segmentation tools (e.g., Jieba\footnote{\url{https://github.com/fxsjy/jieba}}) and input methods (e.g., Sougou\footnote{
\url{https://pinyin.sogou.com}\label{sougou}}), we propose to add an alterable user dictionary to handle the adaption problem of CSC in the zero-shot schema.
Since the output probability in token classification is independent token-by-token, the user dictionary can not be directly adopted in such spellers.
Hence we propose a novel \textbf{U}ser \textbf{D}ictionary guided inference module (UD) to post-process the prediction results of a token classification based speller.
Specifically, given the predicted token path candidates, UD rewards the paths containing more dictionary terms.
Notably, this method applies to both general and domain adaptive CSC tasks.

Due to the lack of existing evaluation datasets, we annotate a CSC dataset of the law, medical treatment and official document writing domains to represent these three scenarios.
We both conduct experiments on the general benchmark \cite{DBLP:conf/acl-sighan/TsengLCC15} and our domain specific evaluation datasets.
Our experimental results show that our ECSpell achieves new state-of-the-art performance on the general benchmark and UD can improve the performance of all tests with no extra fine-tuning.
We combine our ECSpell and UD to form our CSC extensive speller ECSpell$^{UD}$.
Related experiments demonstrate that ECSpell$^{UD}$ surpasses all the other baselines.

Our contributions can be summarized as follows:
\begin{itemize}
\item We develop an error consistent masking strategy for unsupervised CSC learning.
\item We annotate domain specific datasets for follow-up CSC research.
\item We use the user dictionary guided inference module (UD) to enhance the domain adaptive for any token classification based speller.
\end{itemize}

We organize the rest of this paper as follows.
First, in Section 2, we introduce the related work.
Following Section 3 and Section 4, we describe our work for the general CSC task and user dictionary guided domain adaptive CSC task, respectively.
Next, Section 5 presents the experimental result and related analysis.
Finally, we conclude our work in Section 6.

\section{Related Work}~\label{ccc}
\subsection{Chinese Spelling Check}
Previous research on CSC can be divided into three distinct categories:
rule based methods, machine learning based methods and deep learning based methods.
\citet{jiang2012} utilized the China National Matriculation Examinations (NME) rules to classify spelling errors into Idiom Error and Word Error.
\citet{DBLP:conf/acl-sighan/YuL14} used character-level n-gram language models and a word vocab to detect and correct potential misspelled characters.
For machine learning based methods,
CRF-based word segmentation/part of speech tagger was integrated into a tri-gram language model after the rule-based fronted \cite{DBLP:conf/acl-sighan/WangLWC13, gu-etal-2014-introduction}.

With the rapid development of deep learning techniques,
the process of CSC has moved a big step forward.
\citet{wang-etal-2019-confusionset} leveraged the pointer network by picking the correct character from the confusion set.
\citet{hong2019faspell} firstly modeled CSC as a BERT token classification task.
As most spelling errors come from similar pronunciations or glyphs,
many successive studies merged the similarity knowledge into spellers.
\citet{DBLP:journals/taslp/NguyenNC21} adopted glyph features while \citet{DBLP:conf/acl/ChengXCJWWCQ20, zhang-etal-2021-correcting}, and \citet{xu-etal-2021-read} employed phonetic information.
Researcher also explored the mixture method of these similarities, 
such as an adaptive gating module \cite{xu-etal-2021-read}, GCN \cite{DBLP:conf/acl/ChengXCJWWCQ20} and multi-modal \cite{huang-etal-2021-phmospell}.
To refine the learning object, \citet{DBLP:conf/acl/LiZZH20} applied an adversarial strategy to enhance the robustness of the model while \citet{li2021tailtotail} adopted the focal loss penalty strategy to alleviate the class imbalance problem.
Further, \citet{DBLP:conf/acl/LiZLLLSWLCZ22} refined the knowledge representation of pretrained language models to narrow the gap between it and the essential of CSC task via the contrastive learning method.
\citet{zhang-etal-2020-spelling} combined the losses of detection and correction with a soft-mask strategy.
Different from the above mentioned approaches, \citet{DBLP:conf/emnlp/BaoLW20} applied a non-autoregressive model to improve the phrase correction performance.

\subsection{Domain Terminology Injection}
In the development of the neural machine translation (NMT), significant studies proposed methods to integrate external specialized domain terminologies.
It can be roughly divided into the following three categories \cite{DBLP:conf/coling/MichonCS20}:
1). Placeholders.
Placeholders replace terms appearing in the sentence pair in pre- and post-processing \cite{crego2016systran}.
It is obviously that this method lacks flexibility as the model will always replace the placeholder with the same phrase irrespective of grammatical context.
2). Learning to apply constraints.
It learns a copy behavior of terminology at training time \cite{song2019code, dinu2019training}. 
For example, “\begin{CJK}{UTF8}{gbsn}疟疾\textcolor[rgb]{0,0,1}{疟疾}是一种由寄生虫引起的威胁生命的疾病\end{CJK}” (Malaria \textcolor[rgb]{0,0,1}{Malaria} is a life-threatening disease caused by parasites).
The model is trained to incorporate terminology translations when provided as additional input in the source sentence.
It also lacks generalization power as it simply ``copy'' the term found in the terminology base on the source sentence, irrespective of the target hypothesis context \cite{dinu2019training}.
3). Constrained decoding.
The model takes the translation terms as the decoding constraints applied in the inference stage \cite{hokamp2017lexically, hasler2018neural, susanto2020lexically}.
However, a source term may have multiple translation term inflections among which the MT engine should on-the-fly select the best one depending on the source and target context.

As far as we known, there is little study devoted to the domain adaptive spelling check although it is quite important in real life.

\section{General CSC Task}
\subsection{Problem Formulation}
Given an input sentence \(\mathit{X}=\{x_1, x_2, \cdots, x_n\}\) of length \(n\), the model needs to generate its corresponding correct sentence  \( \mathit{Y}=\{y_1, y_2, \cdots, y_n\}\).
For CSC, most tokens in output sentence \(\mathit{Y}\) are the same as that in the input sentence \(\mathit{X}\) and the rest is the target of error correction.
Since the length of input sentence \(\mathit{X}\) is equal to the output sentence \(\mathit{Y}\), this task is usually formed token classification task \cite{hong2019faspell, DBLP:conf/acl/ChengXCJWWCQ20}.

\subsection{Our ECSpell Overview}~\label{ecspell_model}
\begin{figure}
    \centering
    \includegraphics[width=0.8\linewidth]{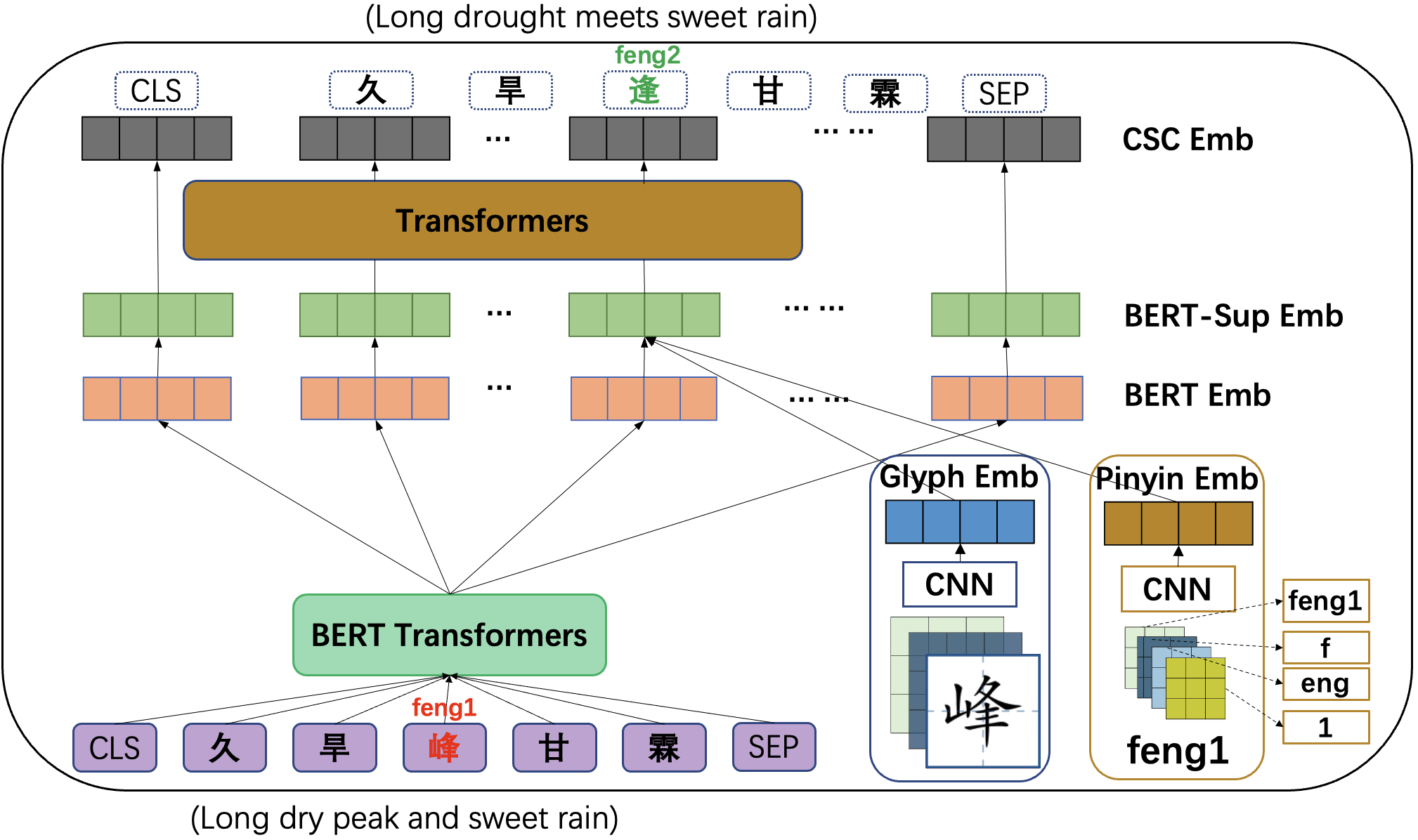}
    \caption{Overview of our general model.}
    \label{fig:model}
\end{figure}
Since spelling errors focus on shape and sound similarities, we enrich the BERT token classification model with related features as shown in Figure \ref{fig:model}.
In order to make more effective use of these two knowledge mentioned above,
our general model ECSpell has made two improvements.
On one hand, the knowledge of glyph including the meaning of radical and the frame structure are considered.
On the other hand, we explore the fusion of pinyin\footnote{\url{https://en.wikipedia.org/wiki/Pinyin}} knowledge.
\subsubsection{Embedding Layer}
\paragraph{\textbf{BERT Embedding}}
Following previous works \cite{xu-etal-2021-read, huang-etal-2021-phmospell, hong2019faspell}, we employ BERT \cite{bert} encoder as the semantic encoder.
Benefit from the pretraining on large-scale corpus, BERT embedding of a sentence contains its rich contextual information.

For the input sentence $X=\{x_1, x_2, \ldots, x_n\}$, the semantic embedding $E^S = \{e_1^S, e_2^S, \ldots, e_n^S\}$ is formulated as follows:
\begin{equation}
    e^S_i = BERTEncoder(x_i)
\end{equation}
where $E^S \in \mathbb{R}^{l\times768}$ and $l$ is the length of the input sentence.

\paragraph{\textbf{Glyph Embedding}}
Compared with recent research, we pay more attention to the radical meaning and the frame structure of Chinese characters.
The Glyce encoder \cite{DBLP:conf/nips/MengWWLNYLHSL19} which utilizes a Tianzige-CNN structure is adopted as the visual information encoder.
To some extent, it fits the origin of Chinese characters better than other methods, such as stroke sequence \cite{liu-etal-2021-plome, DBLP:conf/ialp/HanLWF19} or object detection \cite{huang-etal-2021-phmospell}.
Taking account of the evolution of Chinese characters and their current form, we select two fonts (\begin{CJK}{UTF8}{gbsn}楷书，k\v{a}ish\={u}\end{CJK}) in both traditional and simplified Chinese finally.

Given an input sentence $X=\{x_1, x_2, \ldots, x_n\}$, we define its glyph embedding $E^G = \{e_1^G, e_2^G, \ldots, e_n^S\}$ as follows:
\begin{equation}
    e^{G}_i = GlyphEncoder(x_i)
\end{equation}
where $E^G \in \mathbb{R}^{l \times d_g}$ and $d_g$ is the dimension of glyph embedding.

\paragraph{\textbf{Pinyin Embedding}}
For phonetic errors,
following recent research \cite{huang-etal-2021-phmospell, wang-etal-2021-dynamic}, we use \textit{Pinyin} to represent pronunciation.
Previous works usually treated the pinyin as a whole token \cite{liu-etal-2021-plome, zhang-etal-2021-correcting}, ignoring its internal components of the initial, final, and tone.
Although two pinyin strings are different, they may still be similar in phonetic due to the same initials or finals.
For example, the pinyin string of “\begin{CJK}{UTF8}{gbsn}插\end{CJK} (insert)” and “\begin{CJK}{UTF8}{gbsn}擦\end{CJK} (wipe)” are “ch\={a}” and “c\={a}” respectively.
Thus, we separate the integral pinyin of each character into its initial, final and tone.

When given an input sentence $X = \{x_1, x_2, \ldots, x_n\}$, we first convert each pinyin of character $x$ in $X$ to its corresponding separation form.
Then we concatenate pinyin of each character's integral form and separation form as its pinyin representation.
Finally, we apply a CNN network of width 2 and a max pooling function to extract the phonetic information as:
\begin{align}
    p^S_{i} &= [p^{Initial}_{x_i}\cdot p^{Final}_{x_i}\cdot p^{Tone}_{x_i}] \\
    p_{i} &= [p^{I}_{i} \cdot p^{S}_{i}] \\
    e^P_{i} &= CNN(p_{i})
\end{align}
where [$\cdot$] means concatenate operation between embeddings, $p^I_{i}$ and $p^S_{i}$ is the integral pinyin string and separation pinyin form of $x_i$ respectively, $p_i$ is the final pinyin representation, $e^P_{i} \in \mathbb{R}^{l \times d_p}$ and $d_p$ is the dimension of pinyin embedding.

\subsubsection{Output Layer}
The original BERT embedding integrated with glyph embedding and pinyin embedding is fed into the 2-layer Transformer encoder \cite{DBLP:conf/nips/VaswaniSPUJGKP17}:
\begin{align}
    e_i &= \operatorname{MLP}([e^S_i \cdot e^G_i \cdot e^P_i]) \\
    \tilde{e}_i &= \operatorname{LayerNorm(e_i)} \\
    h_i &= \operatorname{Transformer}(\tilde{e}_i) 
\end{align}
where $\operatorname{MLP}$ is a linear layer, $h_i \in \mathbb{R}^{d_t}$ and $d_t$ is the output dimension of the Transformer encoder.

Following \citet{DBLP:conf/nips/MengWWLNYLHSL19}, we combine the loss of token classification task and glyph classification task as the final training objective.
The training objective $\mathcal{L}$ is given as follows:
\begin{align}
    \mathcal{L}_{glyph} &= -\log{p(z|x)} \\
&=-\log{\operatorname{softmax}(W \times h_{image})} \\
    \mathcal{L}_{csc} &= - \sum^{n}_{i=1}\log{P(\hat{y_i}=y_i|X)} \\
\mathcal{L} &= (1-\lambda)\mathcal{L}_{csc} + \lambda\mathcal{L}_{glyph}
\end{align}
where $z$ is the label of font image $x$, $h_{image}$ is the hidden state from CNNs in the Glyce.
$\hat{y_i}$, $y_i$ are the prediction and label of $X$ respectively.
$\lambda$ controls the trade-off between the CSC classification objective and the auxiliary image classification objective, where $\lambda \in [0, 1]$.

\subsection{Error Consistent Masking Strategy}~\label{ecm}
We propose an Error Consistent Masking (ECM) strategy to create misspelling.
In \citet{zhang-etal-2021-correcting} and \citet{liu-etal-2021-plome}, candidates to replace the original character are independently selected from the confusion set.
For example, “\begin{CJK}{UTF8}{gbsn}报导 (b\`{a}o d\v{a}o)\end{CJK}” (report) could be changed into “\begin{CJK}{UTF8}{gbsn}爆异 (b\`{a}o y\`{i}) (explosive difference)\end{CJK}”, where “\begin{CJK}{UTF8}{gbsn}报 (b\`{a}o)\end{CJK}” (newspaper) and “\begin{CJK}{UTF8}{gbsn}爆 (b\`{a}o)\end{CJK}” (burst) hold the same pronunciation while “\begin{CJK}{UTF8}{gbsn}导 (d\v{a}o)\end{CJK}” (guide) and “\begin{CJK}{UTF8}{gbsn}异 (y\`{i})\end{CJK}” (difference) look similar.

However, people use only one input method when inputting sentences in fact.
Especially under the pinyin input method, people are primarily used to inputting continuous characters.
Previous masking strategies break the consistency of spelling errors in one sentence.

In order to better simulate the real input scene, we simulate the idea of generating candidates by input method based on N-gram method and build an N-gram level confusion set:
\begin{itemize}
    \item Step 1. We collect 2-gram, 3-gram, 4-gram spans from large-scale corpus\footnote{\url{https://github.com/brightmart/nlp_chinese_corpus}\label{wiki}}.
    \item Step 2. We match spans in different gram sets using the single character pinyin confusion set which only includes phonetic-similar candidates.
    \item Step 3. To further robust, we segment large-scale corpus sentences, collecting medium and high-frequency phrases.
    These phrases are converted to pinyin\footnote{\url{https://pypi.org/project/pypinyin/}} and reconverted to characters\footnote{\url{https://github.com/letiantian/Pinyin2Hanzi}} by tools.
    We collect the candidate phrases provided in this process.
\end{itemize}

Therefore we get an N-gram confusion set of which size is 57,363 and it can provide candidates if given a fragment.
Notably, this N-gram confusion set contains not only phrases with similar pinyin but also high frequency input segments with similar voices.
For example, “\begin{CJK}{UTF8}{gbsn}一年 (y\={i} ni\'{a}n)\end{CJK} (one year) - \begin{CJK}{UTF8}{gbsn}意念 (y\`{i} ni\`an) (mind)\end{CJK}”, both of which are phrases and “\begin{CJK}{UTF8}{gbsn}四类 (s\`{i} l\`ei)\end{CJK} (four types) - \begin{CJK}{UTF8}{gbsn}室内 (sh\`{i} n\`ei)\end{CJK} (indoor)”, where the former is a high-frequency input segment, and the latter is a phrase.

In addition to the N-gram confusion set, we divide the prior confusion set of a single character into phonetic-similar confusion set and morphological-similar confusion set, according to the error type annotated in the prior one.
Overall, we get three confusion sets and then do the following operations:
\begin{enumerate}
    \item determining the type of error in a sentence according to 30\% of the pronunciation, 30\% of the shape, 20\% of the random and 20\% of the unchanged, referring to \citet{liu-etal-2010-visually}.
    \item selecting single or continuous characters randomly and replacing them with the same type from confusion set or the N-gram confusion set. Continuous errors are selected only when the error type is sound.
\end{enumerate}

The total number of chosen characters is limited to less than 15\% of the sentence length according to the mask ratio in \citet{bert}.

\section{User Dictionary Guided Domain Adaptive CSC Task}
Most of the existing benchmark SIGHAN \cite{DBLP:conf/acl-sighan/WuLL13, DBLP:conf/acl-sighan/YuLTC14, DBLP:conf/acl-sighan/TsengLCC15} focuses on the general domain.
However, it is common for spellers to function within a specific domain.
Unfortunately, based on our information, there is no domain specific CSC evaluation benchmark.
Hence we annotate three domain specific CSC datasets.
Meanwhile, to extend the CSC task to the practical application, we introduce the zero-shot CSC task which aims to be adaptive to specific domains without extra training data.
Accordingly, we also propose a simple but effective framework to handle this problem with the help of the user dictionary.

\subsection{Domain Specific Dataset Construction}
We collect the raw sentences from publicly available websites, consisting of three typical domains: Law\footnote{\url{http://cail.cipsc.org.cn:2020/}}, medical treatment (Med)\footnote{\url{https://github.com/CBLUEbenchmark/CBLUE}} and official document writing (Odw)\footnote{\url{http://www.gov.cn/}}.
The legal data is composed of both the question stems and options of the multiple-choice questions in the judicial examination.
The medical treatment data consists of sentences in QA pairs from online network consultations.
The official document writing data comprises news, policies and national conditions officially reported by the state.
We use all sentences in the public dataset as candidates for legal and medical treatment domains.
As for the official writing domain, we crawl 1,000 official documents and split their content into single sentences using natural separators such as periods and question marks.

We only retain sentences with more than 5 Chinese characters from these raw data and sample them randomly as the candidates.
Then we enlist five native volunteers to copy the raw sentences and create possible spelling errors in the three domains.
The guideline for manufacturing errors is as follows:
\begin{itemize}
    \item Annotators should fully consider the context and the meaning of the selected characters when making spelling errors.
    \item Apart from single-character errors, continuous phrase-level errors should also be considered.
    \item As SIGHAN15, error characters in each sentence should be no more than 15\%, with a maximum total of 7.
\end{itemize}

\begin{table}[tp]
  \centering
    \begin{tabular}{c|c|c|c|c}
    \toprule
          & Law   & Med & Odw  & SIGHAN15 \\
    \midrule
    \# Error sents/Sents & 1,314/2,460 & 1,699/3,500 & 1,259/2,220 & 542/1,100 \\
    Min.Len   & 12     & 11  & 9 & 5 \\
    Max.Len   & 120   & 127 & 161 & 108 \\
    Avg.Len   & 30.5 & 50.1 & 41.2 & 30.7 \\
    \# Continuous error sents  & 229 & 253 & 265 & 51 \\
    \# Annotators per sent& 5  & 5 & 5 & 1 \\
    PPL & 30.71  & 32.45 & 26.67 & 30.76 \\
    \bottomrule
    \end{tabular}%
  \caption{Statistics of different datasets.}
  \label{tab:ds}%
\end{table}%

Finally, the annotation results are merged and manually proofread.
The dataset holds the following three appealing properties.
\paragraph{\textbf{Reduce Subjective Bias}}
Five annotators label each sentence and the results are then proofread manually to minimize the annotation deviation caused by subjective factors.

\paragraph{\textbf{Reduce Bias of Input Methods}}
To simulate people's input as much as possible,
these annotators are required to use different input methods including Microsoft Pinyin\footnote{\url{https://en.wikipedia.org/wiki/Microsoft_Pinyin_IME}}, Google Pinyin\footnote{\url{https://en.wikipedia.org/wiki/Google_Pinyin}}, Tencent Pinyin\footnote{\url{http://qq.pinyin.cn/}}, Baidu Pinyin\footnote{\url{https://srf.baidu.com/}} and Sougou Handwriting\textsuperscript{\ref{sougou}}.
Meanwhile, we ensure that an annotator keeps one input method to label a sentence, just like the real input scenario. 

\paragraph{\textbf{Continuous Misspelling}}
Continuous spelling errors are also common in real life.
We create this type of misspelling via phrase-level phonetic or visual similar replacement, Internet buzzword replacement and different input intervals in continuous input.

We supplement correct sentences to make the correction rate close to SIGHAN15.
The statistic of this dataset is shown in Table \ref{tab:ds}.
We introduce the SIGHAN15 test dataset to make an intuitive comparison.
Compared with the general domain, we choose some domains with representative terms, which can be seen from the differences in the PPL metric in the above table.
As can be seen, the sentences in Law, Med and Odw datasets are much more than that in the SIGHAN15 test dataset.
The proportion of error sentences in our built datasets is similar to that in the SIGHAN15 test dataset.
The average length of sentences in the Med dataset is 50.1, which is longer than others.
In addition, the number of continuous error sents is more than that in the SIGHAN15, which can better restore the real input scene.

\subsection{User Dictionary Guided Framework}
\begin{figure}[tbp]
\centering
    \includegraphics[width=9cm]{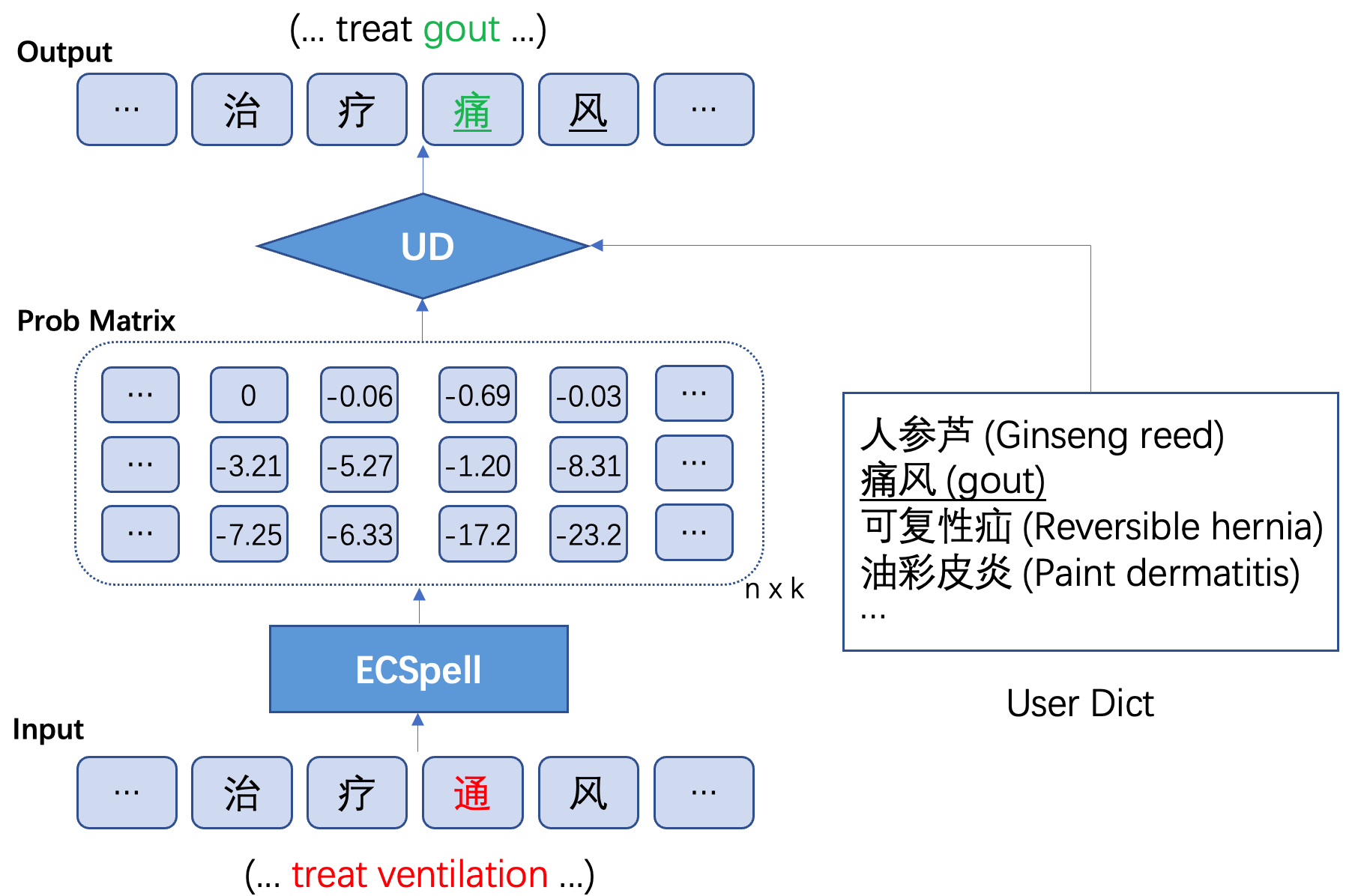}
    \caption{Overview of the UD framework flow chart. Underline characters are included in the user dictionary.}
    \label{fig:ud_flowchart}
\end{figure}

We propose a simple and effective framework named \textbf{U}ser \textbf{D}ictionary (UD) for this domain adaptive CSC task, which is also suitable for classified-based check spellers.
As Figure~\ref{fig:ud_flowchart} shows, UD gives the final prediction according to the probability matrix and user dictionary jointly.
The ECSpell could be replaced with other classified-based check spellers.
Compared with obtaining training corpus of different domains, collecting their corresponding dictionaries saves more time and economy.
In addition, UD can obtain the ability to adapt to various application scenarios by switching user dictionaries.
Our experiment shows that the error correction result will be greatly improved with the help of these dictionaries.

We can first achieve the top-\(k\) rank log-probability matrix \(O\) generated by any token classification based speller.
Then UD will update the prediction result according to the given user dictionary.
Specifically, we can select a token $y^\mathcal{P}_j$ in each column $j$ of $O$ and connect them to form a candidate correction path \(\mathcal{P}=\{y^\mathcal{P}_1,y^\mathcal{P}_2,\cdots,y^\mathcal{P}_n\}\), where $y^\mathcal{P}_j \in O_j$. 
In this way, the score of \(\mathcal{P}\) computed by the original speller is: 
\begin{align} \label{eq:raw_score}
    score_{O}^{\mathcal{P}} = \sum_{j=1}^{n}{score(y^\mathcal{P}_j)}
\end{align}
where $score(y^\mathcal{P}_j)$ is the related log-probability value from $O$.
Since the output probability of token classification based models is independent token-by-token,
the optimal path from Eq.~\ref{eq:raw_score} is simply to pick the first candidate of each position.
Our UD module adopts the following rules to encourage paths including more dictionary terms:

\begin{figure*}[tp]
    \centering
    \includegraphics[width=13.5cm]{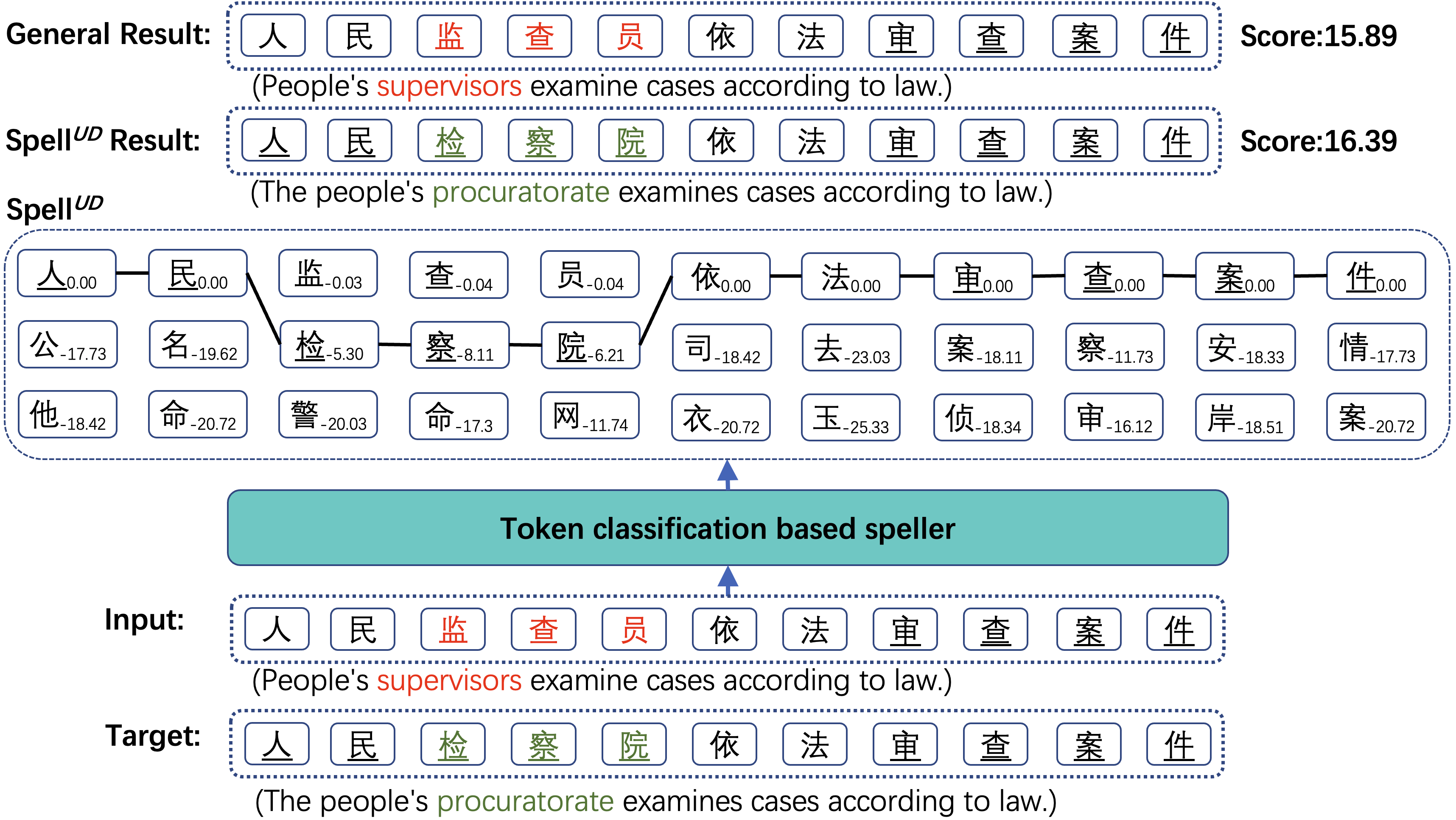}
    \caption{An example of inference process with UD.
    ``\begin{CJK}{UTF8}{gbsn}监查员''\end{CJK} (supervisors) in this sentence should be corrected to ``\begin{CJK}{UTF8}{gbsn}检察院''\end{CJK} (procuratorate).
    Underline characters, ``\begin{CJK}{UTF8}{gbsn}人民检察院” (people's procuratorate) \text{and} “审查案件''\end{CJK} (review cases), are included in the user dictionary.
    The scores of general result and UD inference result are calculated with parameters by Eq.\ref{eq:optimal_path}.}
    \label{fig:inference}
\end{figure*}

\paragraph{\textbf{Raw Span Match (RSM)}}
If the span in the input sentence can be found in the dictionary, we regard this part as correct and keep it unchanged.

\paragraph{\textbf{Altered Span Match (ASM)}}
In the aspect of alternative paths containing dictionary terms, we reward them as follows:
\begin{align}
    score_{D}^{\mathcal{P}} = l_{match}^{\mathcal{P}}
\end{align}
where \( l_{match}^{\mathcal{P}} \) stands for the character number of altered span matched the dictionary in \(\mathcal{P}\).

Finally, the optimal candidate path is selected as follows:
\begin{align}\label{eq:optimal_path}
    \mathcal{P}_{opt} = \mathop{\arg\max}_\mathcal{P}(score_{O}^{\mathcal{P}} +\eta \times  score_{D}^{\mathcal{P}})
\end{align}
where \(\eta\) is a hyper-parameter and we set it to 4 empirically.

For a sentence \(X\) of length \(n\), the total number of candidate paths is \(k^n\).
Besides RSM, we set the minimum and maximal threshold for each token candidate to prune.
On one hand, when the probability of a token candidate exceeds the maximal threshold, we fix that token as the predicted result.
On the other hand, we discard a token if its score is below the minimum threshold.
In the following experiment, the minimum and maximal thresholds are set to -11 and -0.001, respectively.

After pruning, the average numbers of candidate paths in different spellers are shown in Table \ref{tab:avg_candidate}.
We can find that the number varies greatly.
For weak models, there are still too many candidate paths.
It will consume much time if the greedy algorithm obtains the global optimum.
Hence we use beam search to reduce the search space.
To balance the speed and effect, we set the beam size to 20 and the hyperparameter \(k\) to 5.
As can be seen from Table \ref{tab:avg_candidate}, for our model ECSpell\(^{\#}\), it is usually able to select the optimal candidate path while this chance in weaker models like SM BERT and SpellGCN is reduced seriously.

Figure~\ref{fig:inference} shows an example.
The input sentence is ``\begin{CJK}{UTF8}{gbsn}人民监查员依法审查案件\end{CJK}'' (People's supervisors examine cases according to law) where ``\begin{CJK}{UTF8}{gbsn}监查员\end{CJK}'' (supervisors) should be corrected to ``\begin{CJK}{UTF8}{gbsn}检察院\end{CJK}'' (procuratorate).
These potential errors are too confusing to capture by general models.
Nevertheless, UD framework rescores each path according to Eq.~\ref{eq:optimal_path} and then gives the prediction consistent with the target sentence.

\begin{table}[tbp]
  \centering
    \begin{tabular}{l|r|r|r}
    \toprule
     Model     & Law   & Med & Odw\\
          \midrule
    SM BERT & 783.45  & 1258.83 & 106.67 \\
    SpellGCN & 77.29  & 507.04 & 107.85 \\
    BERT$^{\#}$  & 4.50  & 149.85  & 2.89 \\
    ECSpell$^{\#}$ & 3.89  & 38.49 & 2.64 \\
    \bottomrule
    \end{tabular}%
  \caption{The average candidate paths per sentence. "$^{\#}$" denotes the model is additionally pretrained.}
  \label{tab:avg_candidate}%
\end{table}%

\section{Experiments}
\subsection{Data}

\subsubsection{Pretraining}
We collect 38.1 million sentences from news2016$^{\ref{wiki}}$ and wiki2019zh\textsuperscript{\ref{wiki}} for pre-training.
Compared with \citet{liu-etal-2021-plome}, our pre-training is quite lightweight.

\subsubsection{General Task}
We evaluate our general model on the widely-used SIGHAN15 benchmark.
The corpus for fine-tuning consists of the SIGHAN training data (6,476 samples) \cite{DBLP:conf/acl-sighan/WuLL13, DBLP:conf/acl-sighan/YuLTC14, DBLP:conf/acl-sighan/TsengLCC15} and automatically generated pseudo data (271,329 samples) \cite{DBLP:conf/emnlp/WangSLHZ18}.
Follow previous works \cite{hong2019faspell, DBLP:conf/acl/ChengXCJWWCQ20, zhang-etal-2020-spelling, DBLP:conf/acl/LiZZH20},
we convert the traditional Chinese character in SIGHAN dataset to simplified Chinese form using OpenCC\footnote{\url{https://github.com/BYVoid/OpenCC}}.

\subsubsection{Domain Adaptive Task}
Regarding the domain specific CSC evaluation, the dataset we built is employed.
There is no extra data for training.
Any dictionaries can be used to guide the general speller.
In this paper, we adopt the public Tsinghua University open Chinese dictionaries\footnote{\url{http://thuocl.thunlp.org/}} for law and medical treatment domain.
For the official document writing domain, we first crawl 40,000 additional official documents and collect all phrases via the word segment tool.
Then we do the same operations on our pretraining data and obtain general phrases.
Next, we retain the top 10k/15k phrases with the most occurrences in general/Odw phrases as the candidate general/Odw dictionary.
In addition, we adopt the AutoPhrase \cite{shang2018automated} to mine domain focused phrases to supplement original Odw dictionary.
Finally, phrases appearing in the general dictionary are removed from the Odw dictionary.
The statistics related to the dictionary is shown in Table \ref{tab:vocab}.
\begin{table}[tbp]
    \centering
      \begin{tabular}{l|r|r|r}
      \toprule
          & Law & Med & Odw\\
          \midrule
    Num   & 9896 & 18749 & 12509 \\
    Min.Len & 2 & 2 & 2 \\
    Max.Len & 29 & 13 & 14 \\
    Avg.Len & 7.9 & 4.2 & 2.8 \\
    \# Error Related / All & 403/2,460 &1,155/3,500 & 1,158/2,220 \\
    \bottomrule
    \end{tabular}%
  \caption{Statistics of dictionaries of law, medicine and official document writing. 
            “\# Error Related / All”, represents the number of dictionary related error sentences / all sentences in each dataset.}
  \label{tab:vocab}%
\end{table}%
\subsection{Baselines}
\begin{table*}[tp]
  \centering
    \begin{tabular}{l|cccc|cccc}
    \toprule
    \multirow{2}[4]{*}{Method} & \multicolumn{4}{c|}{Detection Level} & \multicolumn{4}{c}{Correction Level} \\
\cmidrule{2-9}          & Acc.  & Pre.  & Rec.  & F1    & Acc.  & Pre.  & Rec.  & F1 \\
    \midrule
    FASPell\(^{\dagger}\) & 67.6  & 67.6  & 60.0  & 63.5  & 66.6  & 66.6  & 69.1  & 62.6  \\
    SM BERT\(^{\dagger}\) & 80.9  & 73.7  & 73.2  & 73.5  & 77.4  & 66.7  & 66.2  & 66.4  \\
    SM BERT &  80.5 & 71.2  & 75.1  &  73.1  & 78.5 &  67.3 & 71.0 & 69.1 \\
    SpellGCN\(^{\dagger}\) & 74.8  & 74.8  & 80.7  & 77.7  & 72.1  & 72.1  & 77.7  & 75.9(74.8) \\
    SpellGCN & 84.0  & 76.9  & 78.9  & 77.9  & 82.7  & 74.4  & 76.3  & 75.4  \\
    \midrule
    DCN-P\(^{\#}\)\(^{\dagger}\) & -     & 77.1  & 80.9  & 79.0  & -     & 74.5  & 78.2  & 76.3  \\
    PLOME\(^{\#}\)\(^{\dagger}\) & -  & 77.4  & 81.5  & 79.4  & -  & 75.3  & 79.3  & 77.2  \\
    REALISE\(^{\#}\)\(^{\dagger}\) & 84.7  & 77.3  & 81.3  & 79.3  & 84.0  & 75.9  & 79.9  & 77.8 \\
    2ways\(^{\#}\)\(^{\dagger}\) & -     & -     & -     & 80.0  & -     & -     & -     & 78.2  \\
    \midrule
    BERT  & 79.4  & 69.8  & 76.6  & 73.0  & 78.4  & 67.9  & 74.5  & 71.1  \\
    BERT\(^{\#}\) & 84.7  & 76.0  & 81.0  & 78.4  & 84.0  & 74.7  & 79.5  & 77.0  \\
    ECSpell & 83.4  & 76.4  & 79.9  & 78.1  & 82.4  & 74.4  & 77.9  & 76.1  \\
    ECSpell\(^{\#}\) & \textbf{86.3 } & \textbf{81.1}  & \textbf{83.0 } & \textbf{81.0 } & \textbf{85.6 } & \textbf{77.5}  & \textbf{81.7 } & \textbf{79.5 } \\
    \bottomrule
    \end{tabular}
  \caption{Performance on the SIGHAN15 test. 
  Best results are in \textbf{bold}.
  "\(^{\#}\)" denotes the model is additionally pretrained.
  "\(^{\dagger}\)" represents the results we quoted.
  Specially, if the precision and recall are correct, F-score of SpellGCN cited in its essay should be 74.8.}
  \label{tab:general}
\end{table*}
\begin{table*}[tp]
  \small
  \centering
    \begin{tabular}{l|cccc|cccc}
    \toprule
    \multirow{2}[4]{*}{Method} & \multicolumn{4}{c|}{Detection Level} & \multicolumn{4}{c}{Correction Level} \\
\cmidrule{2-9}          & Acc.  & Pre.  & Rec.  & F1    & Acc.  & Pre.  & Rec.  & F1 \\
    \midrule
    SpellGCN\(^{\dagger}\) & 83.7  & 85.9  & 80.6  & 83.1  & 82.2  & 85.4  & 77.6  & 81.3 \\
    DCN-P\(^{\#}\)\(^{\dagger}\) & \textbf{94.6}   & 88.0  & 80.2  & 83.9  & 83.2 & 87.6  & 77.3  & 82.1  \\
    PHMOSpell\(^{\#}\)\(^{\dagger}\) & -     & \textbf{90.1 } & 72.7  & 80.5  & -     & \textbf{89.6 } & 69.2  & 78.1  \\
    TtT\(^{\dagger}\)  & 82.7  & 85.4 & 78.1  & 81.6  & 81.5  & 85.0 & 75.6  & 80.0  \\
    \midrule
    ECSpell & 82.9  & 85.7  & 78.4  & 81.9  & 82.0  & 85.4  & 76.6  & 80.7  \\
    ECSpell \(^{\#}\) & 86.3 & 88.3  & \textbf{83.2} & \textbf{85.7} & \textbf{85.6} & 88.1  & \textbf{81.7 } & \textbf{84.8} \\
    \bottomrule
    \end{tabular}
  \caption{Performance on the SIGHAN15 test evaluated by the official tools. 
  Best results are in \textbf{bold}.
  "\(^{\#}\)" denotes the model is additionally pretrained.
  "\(^{\dagger}\)" represents the results we quoted.
  }
  \label{tab:general_official}
\end{table*}
We compare our model with these typical baselines:
\begin{description}
\item[BERT\cite{bert}] The basic BERT token classification model. 
\item[FASPell\cite{hong2019faspell}] This model utilizes a denoising autoencoder to generate candidates from context.
\item[SM BERT\cite{zhang-etal-2020-spelling}] This model uses a soft-masked strategy to combined the detection module and correction module.
\item[SpellGCN\cite{DBLP:conf/acl/ChengXCJWWCQ20}] This model incorporates pronunciation and shape similarity graphs into BERT model via GCN.
\item[DCN\cite{wang-etal-2021-dynamic}] This model considers connection between two adjacent characters.
\item[PLOME\cite{liu-etal-2021-plome}] This model predicts the token and pinyin at the same time.
\item[REALISE\cite{xu-etal-2021-read}] This model adopts a adaptive gate mechanism to fuse the phonetic and visual information.
\item[PHMOSpell\cite{huang-etal-2021-phmospell}] This model integrates pinyin and glyph representations with a multi-modal method.
\item[2ways\cite{DBLP:conf/acl/LiZZH20}] This model uses an adversarial strategy to optimize the robustness.
\item[TtT\cite{li2021tailtotail}] This model adopts focal loss penalty strategy to alleviate the class imbalance problem considering that most of the tokens in a sentence are not changed.
\item[Copy BERT] This model combines a copy behaviour of terminology \cite{dinu2019training} with BERT for token classification method at training time to handle the domain adaptive CSC task.
\end{description}

\subsection{Settings}\label{setting}
Our model is implemented based on huggingface's pytorch implementation of transformers\footnote{\url{https://github.com/huggingface/transformers}}.
Concerning pretraining, the training batch size is 2560 and the training step is 150k, and we utilize AdamW \cite{DBLP:conf/iclr/LoshchilovH19} optimizer with the learning rate of 5e-5.
Instead of training from scratch, we adopt the parameters of Chinese \(\text{BERT}_{wwm}\) to initialize the Transformer blocks.
As to finetuning, the number of fonts in glyce embeddings is 4452 as \citet{DBLP:conf/nips/MengWWLNYLHSL19}.
The training batch size and the learning rate are set to 128 and 2e-5, respectively.
The optimizer is the same as that of pretraining period.
In finetuning, there are two ways of initializing the model parameters.
One is adopting the weight of Chinese \(\text{BERT}_{wwm}\) directly and the other is initialized with the weight obtained from the pretraining period mentioned before.

\subsection{Evaluation Metrics}
To evaluate the performance, we adopt the widely used metrics following \citet{hong2019faspell} and \citet{zhang-etal-2020-spelling}.
Compared with character-level metrics, sentence-level metrics are more rigorous since they measures the ability to detect and correct the spelling errors for the entire sentence.
Metrics includes accuracy, precision, recall and F1.

Besides, we also report the experimental result of general task evaluated by official tool.
In order to facilitate the following research, we also rewritten the original java code version of the official tool into Python version and verify it on all the relevant experiments mentioned in this paper\footnote{The python version code of official evaluation tool is released along with our source code.}.

\subsection{Main Results}\label{main_res}
\paragraph{General Task}
The performance on the SIGHAN15 test set is shown in Table~\ref{tab:general} and Table~\ref{tab:general_official}.
The former is evaluated by the method in \citet{hong2019faspell} while the latter is computed with the official tool.
As can be seen, our model ECSpell$^\#$ outperforms all previous methods.
Table~\ref{tab:general} shows it achieves 2.5\% gain on correction F1 score compared with the basic BERT$^\#$.
From the results, the effect of pretraining for CSC task is very obvious, especially BERT$^\#$.
Even without pretraining, ECSpell can compete against DCN-P$^\#$.
In Table \ref{tab:general_official}, the result is consistent.
Our ECSpell$^\#$ achieves the best performance.
It is noticed that PHMOSpell$^\#$ performs well in precision but not in recall.
We deduce that the multi-modal features they incorporated are too strict to mine the potential errors.
By contrast, ECSpell$^\#$ has a better trade-off between precision and recall, and it achieves better performance in terms of the ultimate correction-level F1 score.
The overall results indicate that our enhanced embedding representation and error consistent masking strategy take effect indeed.
Notably, \citet{DBLP:conf/acl/LiZZH20} applies a complex adversarial training mechanism to largely strengthen the model robustness.
We believe that the ECSpell will be further improved if we use the same training mechanism.

\paragraph{Domain Adaptive Task}
\begin{table*}[tp]
  \centering
  \small
  \resizebox{\linewidth}{!}{
    \begin{tabular}{l|cccc|cccc}
    \toprule
    \multirow{2}[4]{*}{Model} & \multicolumn{4}{c|}{Detection} & \multicolumn{4}{c}{Correction} \\
\cmidrule{2-9}          & Acc.  & Pre.  & Rec.  & F1    & Acc.  & Pre.  & Rec.  & F1 \\
    \midrule
    Copy BERT\(^{\#}\) &    80.2/59.8    &   76.1/72.4    &     65.9/59.8   &   70.6/65.5     &     74.7/52.1       &      64.2/63.1     &     55.6/52.1   & 59.6/57.1\\
    \midrule
    SM BERT &  65.7/41.2    & 53.5 /52.9 & 48.3/41.2  & 50.8/46.3  &  58.5/32.0  & 38.4/41.1  & 34.7/32.0  & 36.5/36.0  \\
    \quad+\ UD &   +0.7/+8.2    & +1.7/+\textbf{8.1}  & +1.0/+8.2  & +1.3/+8.3  & +0.6/+10.7  & +1.4/+\textbf{11.7}  & +0.9/+10.7  & +1.1/+11.2  \\
    \midrule
    SpellGCN & 64.0/35.7  & 53.4/50.2  & 44.0/35.7  & 48.3/41.7  & 54.9/23.6  & 32.6/33.1  & 26.9/23.6  & 29.5/27.5  \\
    \quad+\ UD & +1.4/+7.5   & +1.9/+5.9   & +2.4/+7.5   & +2.1/+7.1   & +1.7/+10.6   & +\textbf{3.0}/+11.4  & +3.0/+10.6   & +\textbf{3.0}/+11.2  \\
    \midrule
    BERT\(^{\#}\) & 80.7/56.8  & 76.9/69.6  & 65.5/56.8  & 70.8/62.6  & 77.1/50.9  & 69.0/62.3  & 58.8/50.9  & 63.5/56.0  \\
    \quad+\ UD & +1.7/+\textbf{9.7}   & +\textbf{2.1}/+6.8   & +\textbf{3.1}/+\textbf{9.7}   & +\textbf{2.6}/+\textbf{8.7}   & +\textbf{1.8}/+\textbf{11.6}   & +2.4/+9.9   & +\textbf{3.2}/+\textbf{11.6}   & +2.8/+\textbf{11.0}  \\
    \midrule
    ECSpell\(^{\#}\) & 80.2/57.1  & 76.5/70.3  & 65.0/57.1  & 70.3/63.0  & 77.4/52.6  & 70.5/64.8  & 59.9/52.6  & 64.8/58.1  \\
    \quad+\ UD & +\textbf{1.9}/+9.5   & +1.7/+5.5   & +2.8/+9.5   & +2.3/+6.4   & +1.6/+10.9   & +1.7/+8.1   & +2.7/+10.9   & +2.4/+9.8  \\
    \bottomrule
    \toprule
    Copy BERT\(^{\#}\) & 72.5/46.8   &    59.6/60.4  &     53.9/46.8     &   56.6/52.7  &     67.9/38.4    &   49.1/49.6   &   44.4/38.4  & 46.7/43.3\\
    \midrule
    SM BERT & 61.5/35.6   & 43.6/49.1  & 41.0/35.6  & 42.3/41.3  &  53.8/20.9 & 26.7/28.6  & 25.1/20.8  & 25.9/24.1  \\
    \quad+\ UD & +0.9/+6.4 & +0.8/+3.4  & +4.1/+6.4  & +2.4/+5.6  & +1.2/+7.4 & +0.1/+3.7  & +4.1/+6.7  & +2.1/+5.6 \\
    \midrule
    SpellGCN & 56.7/25.8  & 35.6/38.2  & 32.7/25.8  & 34.1/30.8  & 48.9/11.6  & 18.2/17.2  & 16.8/11.6  & 17.5/13.9  \\
    \quad+\ UD & +1.3/+6.9   & +0.7/+3.1   & +4.8/+7.0   & +2.8/+5.8   & +1.2/+7.3   & +0.3/+4.1   & +4.0/+7.1   & +2.1/+6.0 \\
    \midrule
    BERT\(^{\#}\) & 79.6/55.0  & 74.5/72.9  & 61.4/55.0  & 67.3/62.7  & 76.0/48.3  & 65.6/64.1  & 54.0/48.3  & 59.2/55.1  \\
    \quad+\ UD & +2.2/+7.1   & +\textbf{1.5}/+5.2   & +\textbf{5.8}/+7.1   & +\textbf{3.7}/+7.6   & +\textbf{1.8}/+8.7   & +\textbf{0.8}/+5.8   & +3.6/+8.7   & +\textbf{3.7}/+7.8  \\
    \midrule
    ECSpell\(^{\#}\) & 79.1/54.2  & 75.2/71.7  & 60.6/54.2  & 67.1/61.7  & 76.0/49.6  & 67.3/65.6  & 54.2/49.6  & 60.0/56.5  \\
    \quad+\ UD & +\textbf{1.9}/+\textbf{9.6}   & +0.6/+\textbf{5.9}   & +5.2/+\textbf{9.6}   & +3.3/+\textbf{8.8}   & +1.5/+\textbf{10.5}   & +0.3/+\textbf{6.6}   & +\textbf{4.4}/+\textbf{10.5}   & +2.8/+\textbf{9.1} \\
    \bottomrule
    \toprule
    Copy BERT\(^{\#}\) &     79.3/67.7    &  78.6/79.9 & 64.9/67.7   &    71.1/73.3   &  74.5/59.4    &   68.3/70.1   &  56.4/59.4     &  61.8/64.3\\
    \midrule
    SM BERT & 62.4/46.3   & 52.1/59.9  & 44.9/46.3  & 48.2/52.2  &  55.0/33.2 & 37.1/42.9  & 31.9/33.2  & 34.3/37.5  \\
    \quad+\ UD & +0.7/+6.8 & +1.3/+3.2  & +4.1/+6.7  & +2.9/+5.4  & +1.1/+7.1 & +0.4/+\textbf{4.3}  & +3.2/+5.5  & +2.0/+5.0 \\
    \midrule
    SpellGCN & 60.6/39.5  & 53.1/59.0  & 38.4/39.5  & 44.6/47.3  & 51.7/23.7  & 31.6/35.4  & 22.8/23.7  & 26.5/28.4  \\
    \quad+\ UD & +0.2/+0.5   & +0.5/+0.6   & +0.5/+0.5   & +0.5/+0.6   & +0.5/+0.8   & +1.0/+0.8   & +0.8/+0.8   & +0.9/+1.0 \\
    \midrule
    BERT\(^{\#}\) & 78.4/64.9  & 79.8/80.0  & 62.6/64.9  & 70.1/71.7  & 75.9/60.7  & 74.0/74.8  & 58.1/60.7  & 65.1/67.0  \\
    \quad+\ UD & +\textbf{4.0}/+\textbf{7.6}   & +\textbf{2.8}/+\textbf{3.4}   & +3.2/+\textbf{7.6}   & +\textbf{6.8}/+\textbf{5.9}   & +\textbf{4.5}/+\textbf{6.7}   & +\textbf{1.9}/+2.7   & +4.0/+\textbf{6.7}   & +3.6/+4.9  \\
    \midrule
    ECSpell\(^{\#}\) & 79.1/66.2  & 81.4/81.6  & 63.6/66.2  & 71.4/73.1  & 76.8/62.5  & 76.3/77.0  & 59.6/62.5  & 67.0/69.0  \\
    \quad+\ UD & +3.4/+7.0   & +1.0/+2.0   & +\textbf{6.5}/+7.0   & +4.4/+5.0   & +3.1/+6.6   & +0.6/+2.7   & +\textbf{4.7}/+6.6   & +\textbf{3.2}/+\textbf{5.2}  \\
    \bottomrule
    \end{tabular}}%
  \caption{Performance on the Law (top), Med (mid) and Odw (bottom) test. "\(^{\#}\)" denotes the results with additional pretraining.
  Best results are in \textbf{bold}.
  The values to the left/right of the slash distinguish the result of the entire dataset and the result of the sentences containing dictionary-related errors.}
  \label{tab:domainres}%
\end{table*}
\begin{table}[tp] 
  \small
  \centering
      \begin{tabular}{l|l}
      \toprule
    Input & \begin{CJK}{UTF8}{gbsn}患者需要按照\underline{\textcolor[rgb]{0,0.5,0}{剂}量}服用甲苯{\color{red}米坐}片。\end{CJK} \\
          & \makecell[l]{Patients need to take toluidine sitting tablets according to the dose.}\\
          \hline
    w/o UD & \begin{CJK}{UTF8}{gbsn}患者需要按照{\color{red}计}量服用甲苯{\color{red}米坐}片。\end{CJK} \\
          & \makecell[l]{Patients need to take toluidine sitting tablets according to the meter.}  \\
          \hline
    w/ \ \ UD &  \begin{CJK}{UTF8}{gbsn}患者需要按照\underline{\textcolor[rgb]{0,0.5,0}{剂}量}服用\underline{甲苯\textcolor[rgb]{0,0.5,0}{\textit{咪唑}}}片。\end{CJK} \\
          & \makecell[l]{Patients need to take mebendazole tablets according to the dose.}  \\
    \bottomrule
    \end{tabular}%
  \caption{A example of the input and output of ECSpell\(^{\#}\) with/without UD.
  We highlight the wrong or correct characters in red or green color.
  The underlined phrases, “\begin{CJK}{UTF8}{gbsn}剂量\end{CJK}” (dose) and “\begin{CJK}{UTF8}{gbsn}甲苯咪唑\end{CJK}” (mebendazole), are in the dictionary.}
  \label{table:ud_example}%
\end{table}%
The upper, middle and lower parts of Table \ref{tab:domainres} correspond to the results on the Law, Med, and Odw datasets, respectively.
As shown in this table, general models worsen to some extent in domain specific scenarios.
Relying on the copy mechanism, Copy BERT obtains a certain ability to retain domain terminology.
However, its lower correction-F1 score denotes that the copy mechanism can only hold terms rather than find potential errors.
With the help of our user dictionary guided inference module UD, the performance of all the tested spellers rises although there is no extra data for fine-tuning.
In the detection-f1 level, ECSpell$^{\#}$ does not comprehensively exceed BERT$^{\#}$.
Nevertheless, in the correction-f1 level, the former performs better.
It illustrates that ECSpell$^{\#}$ has stronger ability for accurate error correction.

It can be seen that UD is good at identifying and correcting term-in-dictionary errors.
Take Table ~\ref{table:ud_example} as an example. 
The dictionary term “\begin{CJK}{UTF8}{gbsn}剂量\end{CJK}” (dose) is wrongly changed without UD while another dictionary term “\begin{CJK}{UTF8}{gbsn}甲苯咪唑\end{CJK}” (mebendazole) can be corrected only with UD.
The experimental results also verify that with the help of shifting user dictionaries of different domains, the model could have a good adaptability to its current applying domain field when checking spelling errors.

Moreover, we find that UD tends to work more effectively and efficiently for stronger general models. 
Stronger general models are more likely to contain terms in candidate paths and the number of these paths would be less than weaker ones as Table \ref{tab:avg_candidate} shows. 
As a result, the search space of such a model is less which benefits its efficiency.

Overall, our ECSpell model performs best among all the tested spellers regardless of whether it is guided by UD or not.
Extraordinarily, ECSpell$^{UD}$ can approach the performance on the general benchmark.

\subsection{Ablation Study} \label{abla}

We explore the contribution of each component in ECSpell$^{UD}$ with pretraining by conducting ablation studies with the following settings:
\begin{table}[tp]
  \small
  \centering
    \begin{tabular}{l|cc|cc|cc}
    \toprule
    \multicolumn{1}{c|}{\multirow{2}[4]{*}{Model}} & \multicolumn{2}{c|}{Law} & \multicolumn{2}{c|}{Med} & \multicolumn{2}{c}{Odw}\\
\cmidrule{2-7}          & D-F1. & C-F1. & D-F1. & C-F1. & D-F1. & C-F1.\\
    \midrule
    ECSpell$^{UD}$ & 72.9  & 67.2  & 70.4  & 62.8 &75.8 & 70.2\\
    \midrule
    \quad w/o\ glyph & 70.8  & 64.3  & 67.7  & 60.1 & 73.7 & 67.7 \\
    \quad w/o\ pinyin & 71.5  & 65.2  & 68.9  & 61.9 & 74.1 & 68.8 \\
    \quad w/o\ inter-c & 72.2  & 65.8  & 69.1  & 62.2 & 75.0 & 69.6 \\
    \quad w/o\ ECM & 71.3  & 65.1 & 68.5  & 61.3 & 73.5 & 68.5 \\
    \midrule
    \quad w/o\ RSM   & 72.8  & 67.0  & 69.1  & 62.4 & 75.5 & 70.0 \\
    \quad w/o\ ASM   & 70.2  & 63.9  & 67.5  & 60.7 & 71.6 & 67.2\\
    \bottomrule
    \end{tabular}%
  \caption{Ablation results on our built dataset.}
  \label{tab:ablation}
\end{table}%
1) removing the glyph information,
2) removing the pinyin information,
3) removing internal components of pinyin (denoted as inter-c),
4) removing the error consistent masking (ECM) strategy,
5) removing the raw span match (RSM) rule in UD,
6) removing the altered span match (ASM) rule in UD.

The result can be seen in Table \ref{tab:ablation}.
For general models, removing glyph, pinyin or internal components of pinyin leads to performance degradation.
This proves that the glyph information, as well as the entire pinyin information, i.e., pinyin with its initial, final and tone, could benefit CSC.
Then we remove the consistent masking strategy.
The drop in performance indicates the importance of this strategy.
In order to further study the role of two modules in UD, we remove RSM and ASM, respectively.
The decline in experimental results illustrates that they are both effective while ASM is more significant.
This phenomenon is also very logical.
Although RSM allows UD to reduce the probability of incorrect modification of some terms, the rule is too strict that if a fragment containing spelling errors matches, these errors will persist.
By contrast, ASM makes the error correction process more fault tolerant.
UD would synthesize the prediction results according to the probability predicted by the model and the user dictionary.
To sum up, we can observe that removing any components in ECSpell$^{UD}$ with pretraining brings a performance decline.

\subsection{Upper Bound Analysis}
\begin{figure}[ht]
\centering
    \includegraphics[width=7.2cm]{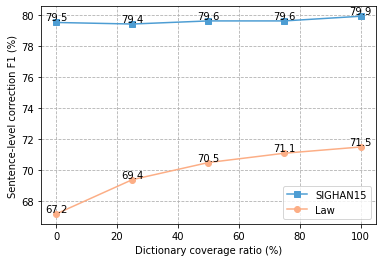}
    \caption{The upper bound analysis results.}
    \label{fig:upper}
\end{figure}
We attempt to build ideal dictionaries consisting of different proportions of original phrase-level errors in each dataset.
Specifically, we use the word segmentation tool to split original error sentences in each dataset.
Then we collect different proportions of error phrases from the word segmentation results to form the ideal dictionaries.
We use the ideal dictionaries to verify the ability of UD to update in real time according to the dictionary.
At the same time, it reflects the upper bound of UD to a certain extent.
Figure \ref{fig:upper} demonstrates that the performance has been further improved with the increase of dictionary coverage ratio.
However, gradually slow growth and the final ceiling indicate that the performance of UD relies on the quality of the dictionary as well.
For example, if there are several conflicting terms which is hard to distinguish in the target domain like “\begin{CJK}{UTF8}{gbsn}检察 (ji\v{a}n ch\'{a})\end{CJK}” (procurator).
Furthermore, “\begin{CJK}{UTF8}{gbsn}监察 (ji\={a}n ch\'{a})\end{CJK}” (monitor), the effect of our UD will also be limited.
Meanwhile, it is also challenging to deal with the independent character error such as “\begin{CJK}{UTF8}{gbsn}他\text{,}\ 她,\ 它\end{CJK} (t\={a}, t\={a}, t\={a})” (he, she, it).

Overall, the proposed method has a particular ability to improve performance in general and domain datasets with a more specific user dictionary.

\section{Conclusion}
In this paper, we developed an error consistent strategy for unsupervised CSC learning and annotated domain specific datasets.
Meanwhile, we proposed a domain adaptive speller ECSpell$^{UD}$ which is composed of a competitive general speller ECSpell and a generic user dictionary guided inference module UD.
Experimental results showed that the performance of ECSpell$^{UD}$ on domain-related benchmarks can approach the level of general benchmarks.

We believe our work can be extended in various aspects.
On one hand, it is meaningful to collect spelling errors from more different domains especially the emerging areas containing the latest terms.
On the other hand, in addition to the domain adaption problem, our model is likely to fit the personalized scenarios.

\begin{acks}
  We thank all reviewers for their valuable comments.
  This work was supported by the Young Scientists Fund of the National Natural Science Foundation of China (No. 62106165), the National Natural Science Foundation of China (No. 62076173, U1836222), the High-level Entrepreneurship and Innovation Plan of Jiangsu Province (No. JSSCRC2021524), and the Project Funded by the Priority Academic Program Development of Jiangsu Higher Education Institutions.
\end{acks}

\bibliographystyle{ACM-Reference-Format}
\bibliography{sample-base}

\appendix

\end{document}